%% file: main.tex
\documentclass[10pt,twocolumn,letterpaper]{article}
\usepackage[pagenumbers]{cvpr}
\input{preamble}
\usepackage{pifont}
\providecommand{\reqyes}{\ding{51}}
\providecommand{\reqno}{\ding{55}}

\definecolor{cvprblue}{rgb}{0.21,0.49,0.74}
\usepackage[pagebackref,breaklinks,colorlinks,allcolors=cvprblue]{hyperref}
\usepackage[capitalize]{cleveref}
\hypersetup{pdftitle={Looking Within: Training-Free Zero-Shot Point Cloud Anomaly Detection via Affine Self-Modeling},pdfauthor={Zhiqiang Qin; He Xie; Junfei Yi; Yunkang Cao; Yaonan Wang; Hui Zhang}}
\title{Looking Within: Training-Free Zero-Shot Point Cloud\\Anomaly Detection via Affine Self-Modeling}
\author{Zhiqiang Qin\textsuperscript{1}, He Xie\textsuperscript{1}\thanks{Corresponding author: He Xie.}, Junfei Yi\textsuperscript{1}, Yunkang Cao\textsuperscript{1}, Yaonan Wang\textsuperscript{1}, and Hui Zhang\textsuperscript{1}\\[-2pt]
{\normalsize \textsuperscript{1} College of Artificial Intelligence and Robotics, Hunan University, Changsha, China}}
\begin{document}
\maketitle
\captionsetup[table]{skip=4pt}
\captionsetup[figure]{skip=8pt}
\flushbottom
\setlength{\textfloatsep}{3pt}
\setlength{\dbltextfloatsep}{6pt}
\setlength{\floatsep}{10pt}
\setlength{\dblfloatsep}{6pt}
\newlength{\ASMFloatBodyGapTarget}
\setlength{\ASMFloatBodyGapTarget}{16pt}
\newcommand{\ASMFloatBodyGap}[1]{\vspace{\dimexpr\ASMFloatBodyGapTarget-#1\relax}}
\begin{abstract}
Point cloud anomaly detection aims to identify geometric defects in three-dimensional objects.
Detecting anomalies from a single test point cloud remains challenging when normal reference objects, additional test instances, and text prompts are unavailable.
We propose \textbf{Affine Self-Modeling (ASM)}, a training-free zero-shot method for point cloud anomaly detection.
{ASM aims to detect anomalies by examining the object's own properties.}
{In feature space, we construct a query-specific low-rank affine model from the mean and dominant variation directions of similar regions within the same object.}
{The affine residual measures each query's departure from this locally supported variation, providing scores for point-level localization and object-level detection.}
{ASM enables efficient single-object inference without task-specific training or external references.}
Extensive experiments on Real3D-AD, Anomaly-ShapeNet, and IEC3D-AD demonstrate that ASM achieves superior performance in \mbox{training-free 3D anomaly detection.}
\end{abstract}
\section{Introduction}
\label{sec:introduction}

\begin{figure}[t]
  \centering
  \includegraphics[width=0.951\columnwidth]{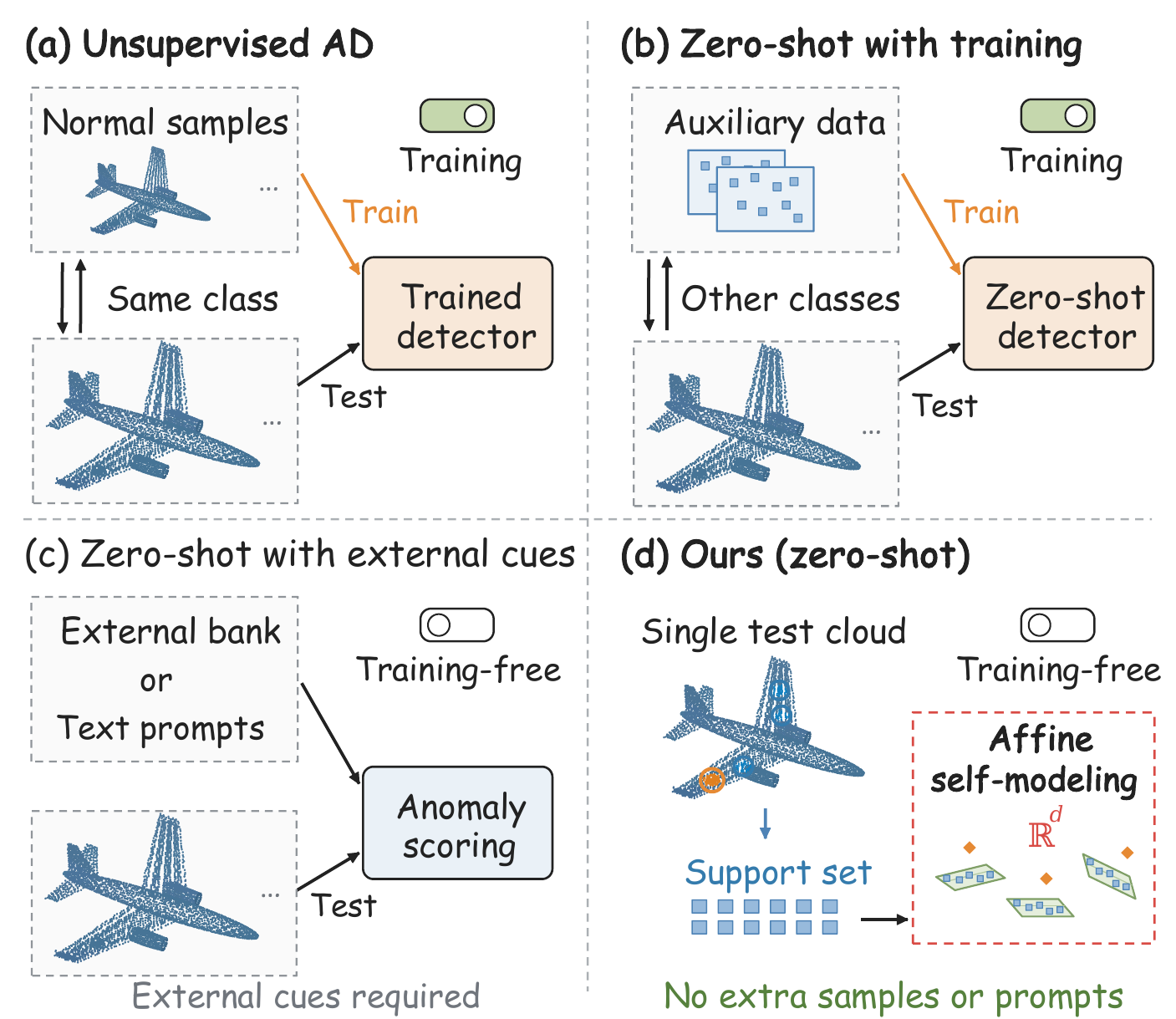}
  \caption{\textbf{Comparison of representative point cloud anomaly detection paradigms.}
  Representative paradigms use (a) normal target-category samples, (b) anomaly cues learned through auxiliary-category training, or (c) external detection cues under training-free inference.
  (d) \method builds query-specific affine models inside one test cloud using a frozen pretrained encoder.
  Planes illustrate low-rank feature-space models. }
  \label{fig:looking-within-teaser}
\ASMFloatBodyGap{6.45pt}
\end{figure}

Point cloud anomaly detection identifies and localizes geometric defects.
Many methods build feature memories or train detectors using normal target-category samples~\citep{ref22,ref32} (Fig.~\ref{fig:looking-within-teaser}(a)).
Such data can be scarce in small-batch or customized production and for newly introduced categories~\citep{sonar2026}.
Zero-shot point cloud anomaly detection addresses this scarcity by detecting anomalies without target-category training samples, reducing reliance on a \mbox{pre-established normal reference set.} Without these training samples, detectors still need cues to distinguish normal from anomalous regions.
Where these cues come from, and what their acquisition and use require, determine the modeling strategy and its \mbox{training or input dependencies.}

Existing zero-shot methods obtain anomaly cues through different routes.
PointAD~\citep{ref42} learns anomaly-related prompts from auxiliary categories (Fig.~\ref{fig:looking-within-teaser}(b)), removing target-category training requirements but retaining task-specific optimization.
Training-free methods avoid downstream optimization (Fig.~\ref{fig:looking-within-teaser}(c)), yet ZUMA~\citep{ref24} combines within-instance statistics with predefined text prompts, and MuSc-V2~\citep{ref19} uses {other unlabeled instances} without text prompts.
We therefore ask whether a single test cloud can provide anomaly cues without task-specific training, normal references, {other instances}, or text prompts.

One possible source lies in internal structural relationships.
For example, a person can notice a chipped gear tooth by comparing it with the remaining teeth, without a separate normal gear.
{This suggests} that the object itself can provide structural cues for anomaly judgment.
{Motivated by this intuition, we turn to \emph{Looking Within} (Fig.~\ref{fig:looking-within-teaser}(d)) to derive modeling support from regions within the same object. Using this support requires a model of the variation among similar regions, against which departures can be assessed.}

{To this end, we propose} \textbf{Affine Self-Modeling (ASM)}, which selects similar intra-object regions for each query using features from a frozen pretrained point encoder.
Their mean and dominant variation directions define a low-rank affine model that accommodates variation along these directions.
Affine residuals measure departures from this local pattern, helping distinguish normal variation from anomalous deviations.
Residual norms support point-level localization and object-level detection.
Each object is processed independently without task-specific \mbox{training or external references.}

We evaluate ASM on Real3D-AD~\citep{ref22}, Anomaly-ShapeNet~\citep{ref17}, and IEC3D-AD~\citep{ref12}.
{ASM achieves point-level AUROC scores of 84.15\%, 91.06\%, and 92.37\%, respectively, and enables efficient and accurate single-object inference in 9.91\,ms (Table~\ref{tab:deployment-speed}).}
Support ablations and matched scoring comparisons validate intra-object support and affine modeling, while cross-encoder experiments show consistent AUROC gains over squared local-mean scoring.
Spatial response visualizations and full-test statistics further show that anomalous regions generally retain a higher residual-energy fraction, consistent with modeling structural variation while retaining anomalous deviations.

We make the following contributions:
\begin{itemize}
\item We introduce ASM for training-free zero-shot point cloud anomaly detection, using a single test cloud for local modeling without normal references, other test \mbox{instances, or text prompts.}

\item We develop query-specific affine residual scoring, using means and dominant variation directions of similar intra-object regions to model local variation. Small neighborhood Gram \mbox{matrices enable efficient computation.}

\item Experiments on three benchmarks validate ASM's effectiveness and efficiency. Across the tested frozen encoders, ASM consistently improves \mbox{point- and object-level AUROC.}

\end{itemize}
\begin{figure*}[t]
  \centering
  \includegraphics[width=0.956\textwidth]{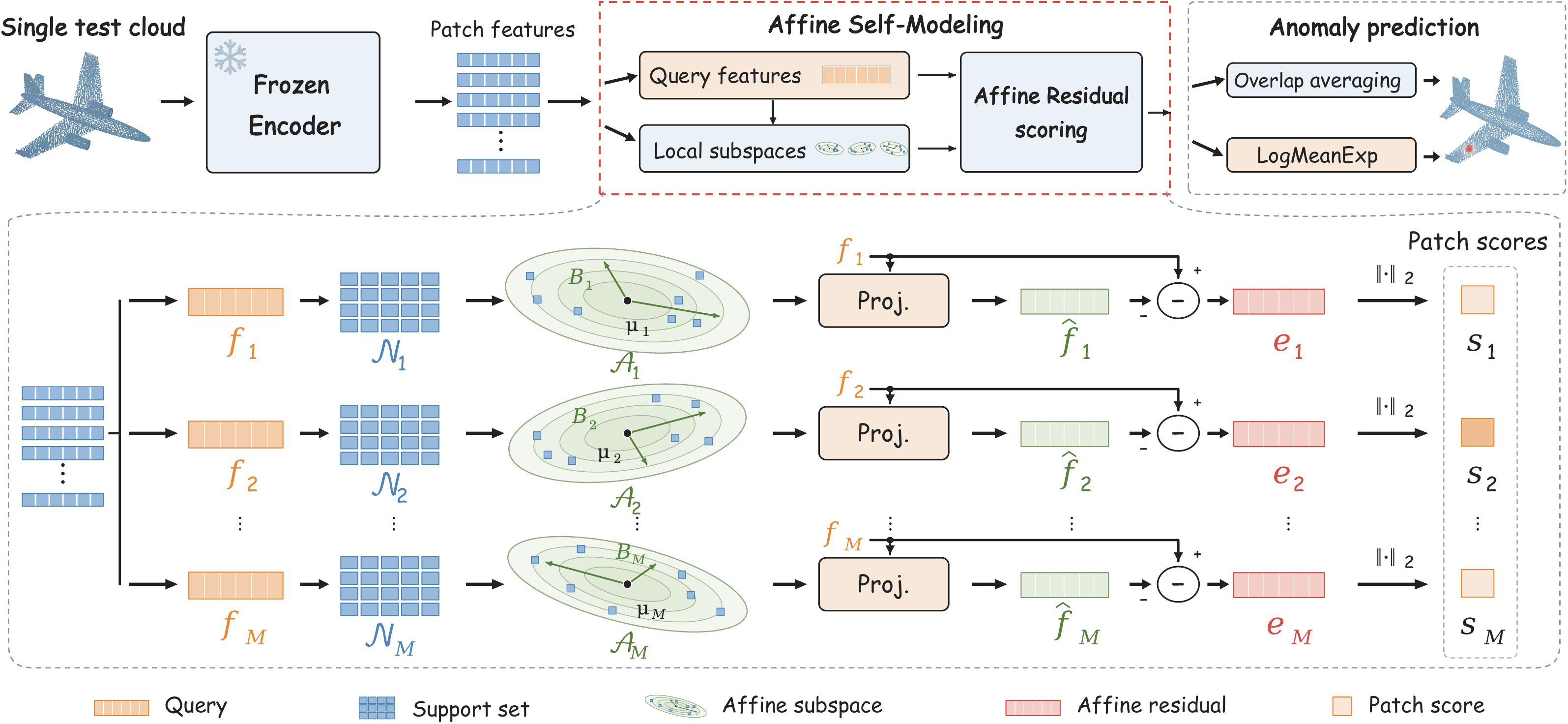}
  \caption{\textbf{Overview of the Affine Self-Modeling (ASM) pipeline.} The top pipeline extracts frozen patch features $\mathbf F$ and produces anomaly predictions. The lower rows detail support selection, affine projection, residual computation, and scoring for queries $\mathbf f_1,\mathbf f_2,\ldots,\mathbf f_M$. Each score is $s_i=\|\mathbf f_i-\widehat{\mathbf f}_i\|_2$. The score vector $\mathbf s$ feeds overlap averaging for point localization and LogMeanExp for object detection.}
  \label{fig:looking-within-workflow}
\ASMFloatBodyGap{4.77pt}
\end{figure*}
\section{Related Work}

\subsection{Point Cloud Anomaly Detection}
Point cloud anomaly detection methods using normal target-category samples focus on geometric description and distillation~\citep{ref03,ref13}, joint geometry and appearance modeling~\citep{ref05,ref08,ref32}, and reconstruction and discrimination~\citep{ref07,ref43}. Subsequent work explores feature contrast, point-offset prediction, and internal spatial modalities to improve 3D \mbox{defect representation and detection~\citep{ref21,ref36,ref44}}.
{Zero-shot methods remove target-category training requirements through transferable representations or auxiliary learning.} PointAD~\citep{ref42}, 3DzAL~\citep{ref33}, and BTP~\citep{ref16} use auxiliary-category prompt learning, synthetic-defect transfer, and point and text alignment, respectively. Related image methods use CLIP representations for text-guided anomaly detection~\citep{ref06,ref14,ref20,ref29,ref39}. For training-free detection, ZUMA~\citep{ref24} combines rendered point-cloud views, text semantics, and within-instance statistics. MuSc~\citep{ref18} and MuSc-V2~\citep{ref19} score unlabeled samples mutually, with the latter supporting both 2D and 3D inputs. 

Other methods detect anomalies in single point clouds by modeling smooth or piecewise-smooth surfaces~\citep{pointsgrade2025,sonar2026}. Their applicability is constrained by assumptions on surface smoothness, curvature, and anomaly sparsity. ASM instead builds query-specific local models in frozen point features and scores each region against the variation learned from \mbox{its own local support.}

\subsection{Subspace and Intra-Instance Modeling}
Anomaly modeling involves both model form and support source. Regarding model form, PCA and its probabilistic formulation~\citep{ref31} capture feature variation in low-dimensional subspaces. SubspaceAD~\citep{ref15} estimates a shared centered PCA model from a few normal images. Its batched zero-shot variant uses all unlabeled test images in a category to build the model and measure deviations of test features.

Regarding support source, intra-instance methods use information within the current test input. Texture anomaly detection uses within-image feature correspondences or neighborhood regression for scoring~\citep{ref01,ref02,grnr2025}. INP-Former~\citep{ref23} extracts intrinsic normal prototypes from test images and detects anomalies through prototype-guided feature reconstruction errors, but training is still required for its \mbox{prototype extractor and decoder.}
ASM combines intra-instance support with local subspace modeling, building a low-rank affine model for each query in the frozen features of the current point cloud and scoring with the affine residual norm using object-local information.
\section{Method}
\label{sec:method}
Figure~\ref{fig:looking-within-workflow} shows the overall pipeline above and query-specific computations below. Each query selects support from the same object, forms an affine model, and obtains its score from the affine residual.

 \subsection{Problem Formulation}
\paragraph{Single-Object Training-Free Zero-Shot Setting.}
As shown in Fig.~\ref{fig:looking-within-teaser}(d), we formulate a new single-object setting for point cloud anomaly detection. A test object is analyzed without anomaly-specific training or external cues. Let $\Phi_{\theta_0}$ be a frozen generic point encoder and $\mathcal P$ the current test cloud. The~detector~is~written~as
\begin{equation}
\begin{aligned}
\mathcal P&=\{\mathbf p_n\}_{n=1}^{N},\quad \mathbf p_n\in\mathbb R^3,\\
\Psi_{\Phi_{\theta_0}}(\mathcal P)&=(\mathbf a,S_{\mathrm{obj}}),\quad
\mathbf a\in\mathbb R^N,\quad S_{\mathrm{obj}}\in\mathbb R.
\end{aligned}
\end{equation}
Here, $\mathbf a=[a_1,\ldots,a_N]^\top$ contains point-level anomaly scores and $S_{\mathrm{obj}}$ is the object-level anomaly score. The setting uses no target-category normal or anomalous training samples, performs no anomaly-specific parameter learning, fine-tuning, or test-time adaptation, and accesses no normal reference objects, additional test instances, or text prompts at inference. Object-specific statistics, support sets, and local models may be estimated from $\mathcal P$, but they are discarded after the current object is processed.

\subsection{Patch Features}
After centering and maximum-radius normalization, we construct $M$ overlapping patches with member indices $\mathcal J_i$. The frozen encoder $\Phi_{\theta_0}$ returns one feature per patch, forming the shared feature matrix in Fig.~\ref{fig:looking-within-workflow}:
\begin{equation}
\mathbf F=\Phi_{\theta_0}(\mathcal P)
=[\mathbf f_1,\ldots,\mathbf f_M]^\top\in\mathbb R^{M\times d}.
\end{equation}
Here, $\mathbf f_i$ denotes the final local token after LayerNorm and FP32 $\ell_2$ normalization. The main configuration uses Point-BERT~\citep{ref38} pretrained with ULIP-2~\citep{ref35}, which extends multimodal alignment in ULIP~\citep{ref34}, with $N=2048$ points, $M=512$ patches, 32 members per patch, and $d=384$.

Transformer interactions preserve patch correspondence while adding context. Scoring ablations share these features, patch grouping, and score aggregation rules.

\subsection{Intra-Object Support Construction}
\label{sec:affine-model}
Each feature $\mathbf f_i$ serves as a query. Its $K$ nearest feature neighbors under distance $d_{\mathrm{ref}}$ define the support indices:
\begin{equation}
\mathcal N_i=\operatorname{KNN}_{K}
(\mathbf f_i,\mathbf F\setminus\{\mathbf f_i\},d_{\mathrm{ref}}).
\end{equation}
Support excludes the query token and may include spatially separated regions. Support is selected from the remaining patches of the same object using feature similarity, without anomaly labels. The query token is excluded, while patch overlap follows the grouping in Sec. 3.2.

\subsection{Local Affine Modeling}
The support mean and centered feature matrix are:
\begin{equation}
\begin{aligned}
\boldsymbol\mu_i&=\frac{1}{K}\sum_{j\in\mathcal N_i}\mathbf f_j,\\
\mathbf H_i&=\begin{bmatrix}
(\mathbf f_{j_1}-\boldsymbol\mu_i)^\top\\
\vdots\\
(\mathbf f_{j_K}-\boldsymbol\mu_i)^\top
\end{bmatrix}\in\mathbb R^{K\times d}.
\end{aligned}
\end{equation}
Local PCA of $\mathbf H_i$ provides orthonormal directions $\mathbf B_i\in\mathbb R^{d\times r_i}$. Together with $\boldsymbol\mu_i$, they define the affine model illustrated in each row:
\begin{equation}
\mathcal A_i=\{\boldsymbol\mu_i+\mathbf B_i\mathbf z
\mid\mathbf z\in\mathbb R^{r_i}\}.
\end{equation}
The exact positive-rank model retains $r_i\leq\min(r,K-1,d)$ directions. Defaults are $K=16$ and $r=8$, with a separate model $\mathcal A_i$ for each query.

\paragraph{Affine Residual Scoring.}
For the centered query $\mathbf v_i=\mathbf f_i-\boldsymbol\mu_i$, orthogonal projection gives:
\begin{equation}
\mathbf v_i=
\underbrace{\mathbf B_i\mathbf B_i^\top\mathbf v_i}_{\mathbf v_i^{\parallel}}
+\underbrace{(\mathbf I-\mathbf B_i\mathbf B_i^\top)\mathbf v_i}_{\mathbf v_i^{\perp}}.
\end{equation}
The parallel component represents model-explained variation. The affine residual $\mathbf e_i=\mathbf v_i^{\perp}$ measures deviation from the orthogonal projection onto $\mathcal A_i$. Its unsquared norm defines the patch anomaly score:
\begin{equation}
 s_i^{\mathrm{ASM}}=\|\mathbf e_i\|_2
 =\left\|(\mathbf I-\mathbf B_i\mathbf B_i^\top)
 (\mathbf f_i-\boldsymbol\mu_i)\right\|_2.
\label{eq:asm-score}
\end{equation}
The projection block in Fig.~\ref{fig:looking-within-workflow} computes $\widehat{\mathbf f}_i=\boldsymbol\mu_i+\mathbf B_i\mathbf B_i^\top\mathbf v_i$. Subtraction gives $\mathbf e_i=\mathbf f_i-\widehat{\mathbf f}_i$, with larger residual norms indicating stronger deviations from the local model.

\paragraph{Relation to Feature Distance.}
Let $\sigma_i^2=K^{-1}\sum_{j\in\mathcal N_i}\|\mathbf f_j-\boldsymbol\mu_i\|_2^2$ denote support dispersion. The mean squared neighbor distance decomposes as follows:
\begin{equation}
\begin{aligned}
D_i&=\frac{1}{K}\sum_{j\in\mathcal N_i}\|\mathbf f_i-\mathbf f_j\|_2^2\\
&=\sigma_i^2+\|\mathbf v_i\|_2^2\\
&=\sigma_i^2+\|\mathbf B_i^\top\mathbf v_i\|_2^2
 +(s_i^{\mathrm{ASM}})^2.
\end{aligned}
\label{eq:distance-decomposition}
\end{equation}
Total distance therefore combines support dispersion, model-explained displacement, and affine residual energy. \method scores only the last component. Two equally distant queries can receive different scores depending on the direction of their displacement. At rank zero ($r_i=0$), the squared score reduces to squared distance to the local mean. Retaining directions extends the mean-only model to account for variation observed in the support.

\paragraph{Efficient Dual-Space Evaluation.}
Since $K\ll d$, we diagonalize the neighborhood Gram matrix:
\begin{equation}
\mathbf G_i=\mathbf H_i\mathbf H_i^\top
=\mathbf U_i\boldsymbol\Lambda_i\mathbf U_i^\top
\in\mathbb R^{K\times K}.
\end{equation}
The implementation retains the top $r$ eigenpairs and uses $\widetilde\lambda_{i,j}=\max(\lambda_{i,j},\epsilon)$, with $\epsilon=10^{-8}$, for inverse square roots. The stabilized score is computed as follows:
\begin{equation}
\widetilde s_i=\left[\|\mathbf v_i\|_2^2-
\left\|\widetilde{\boldsymbol\Lambda}_{i,r}^{-1/2}
\mathbf U_{i,r}^\top\mathbf H_i\mathbf v_i\right\|_2^2
\right]_{+}^{1/2}.
\label{eq:dual-score}
\end{equation}
Here $[x]_+=\max(x,0)$ clips negative residual energies caused by roundoff. Equation~\eqref{eq:dual-score} equals Eq.~\eqref{eq:asm-score} when all retained eigenvalues are at least $\epsilon$ and gives a stabilized approximation otherwise. The derivation is provided in the supplement. Below, $s_i$ denotes the implemented patch score.

\subsection{Anomaly Score}
The patch scores form $\mathbf s=[s_1,\ldots,s_M]^\top$. For each point with coverage count $c_n=\sum_i\mathbb I[n\in\mathcal J_i]$, the overlap-averaging branch computes its anomaly score:
\begin{equation}
 a_n=\frac{1}{c_n}\sum_{i=1}^{M}
 \mathbb I[n\in\mathcal J_i]s_i,\qquad c_n>0.
\label{eq:point-anomaly-score}
\end{equation}
For object-level detection, we use normalized temperature-scaled LogMeanExp~\citep{ref09}, defined as follows:
\begin{equation}
S_{\mathrm{obj}}=\tau\log\left(
\frac{1}{M}\sum_{i=1}^{M}\exp(s_i/\tau)\right).
\label{eq:object-anomaly-score}
\end{equation}
LogMeanExp is permutation invariant, emphasizes high scores as $\tau$ decreases, and preserves constant inputs through $1/M$. We fix $\tau=0.10$ and use identical aggregation rules across all scoring ablations.

\begin{table*}[!t]
\TableRankingColorsOnly
\centering
\caption{Point-level anomaly localization and object-level anomaly detection on Anomaly-ShapeNet and Real3D-AD (\%).}
\label{tab:benchmark-combined}
\label{tab:benchmark-point}
\label{tab:benchmark-object}
\label{tab:benchmark-shapenet}
\label{tab:benchmark-real3d}
\small
\ASMTableSpacing
\renewcommand{\arraystretch}{1.05}
\setlength{\arrayrulewidth}{0.4pt}
\begin{tabularx}{\textwidth}{@{}c|l|>{\hsize=1.16\hsize\linewidth=\hsize\centering\arraybackslash}X*{2}{>{\hsize=.92\hsize\linewidth=\hsize\centering\arraybackslash}X}|>{\hsize=1.16\hsize\linewidth=\hsize\centering\arraybackslash}X*{2}{>{\hsize=.92\hsize\linewidth=\hsize\centering\arraybackslash}X}|cc|>{\centering\arraybackslash}p{21mm}@{}}
\noalign{\hrule height 1.2pt}
\multirow{2}{*}{Dataset} & \multirow{2}{*}{Method} & \multicolumn{3}{c|}{Point-Level} & \multicolumn{3}{c|}{Object-Level} & \multirow{2}{*}{{\makecell{Training\\Free}}} & \multirow{2}{*}{{\makecell{No Ext.\\Refs.}}} & \multirow{2}{*}{Setting} \\
\cline{3-8}
& & P-AUROC & P-AP & PRO & O-AUROC & O-AP & O-F1 & & & \\
\hline
\multirow{9}{*}{\rotatebox[origin=c]{90}{Anomaly-ShapeNet~\citep{ref17}}} & BTF~\citep{ref13}$^{\dagger}$ & 51.00 & 1.80 & 15.30 & 51.00 & 58.30 & 72.20 & {\reqyes} & {\reqno} & \multirow{5}{*}{\textit{Unsupervised}} \\
 & M3DM (MAE)~\citep{ref32}$^{\dagger}$ & 65.10 & 2.90 & 26.60 & 55.10 & 62.00 & 73.10 & {\reqyes} & {\reqno} &  \\
 & PC (MAE)~\citep{ref30}$^{\dagger}$ & 64.50 & 2.80 & 22.10 & 54.70 & 61.80 & 72.50 & {\reqyes} & {\reqno} &  \\
 & CPMF~\citep{ref05}$^{\dagger}$ & 57.30 & -- & -- & 55.90 & 59.70 & -- & {\reqyes} & {\reqno} &  \\
 & Reg3DAD~\citep{ref22}$^{\dagger}$ & 63.20 & 2.40 & 21.90 & 52.70 & 60.40 & 72.00 & {\reqyes} & {\reqno} &  \\
\cline{2-11}
 & BTP~\citep{ref16}$^{\ddagger}$ & \second{78.86} & \second{17.46} & \second{47.43} & \second{65.22} & \second{71.49} & \second{74.86} & {\reqno} & {\reqno} & \textit{ZS} \\
\cline{2-11}
 & ZUMA~\citep{ref24}$^{\ddagger}$ & 65.59 & 3.88 & 31.29 & 57.54 & 64.45 & 72.71 & {\reqyes} & {\reqno} & \multirow{2}{*}{\textit{TFZS~w/~Ext.}} \\
 & MuSc-V2~\citep{ref19}$^{\ddagger}$ & 60.99 & 2.93 & 20.14 & 40.16 & 50.96 & 71.05 & {\reqyes} & {\reqno} &  \\
\cline{2-11}
 & \textbf{ASM}$^{\ddagger}$ & \best{91.06} & \best{31.65} & \best{70.51} & \best{71.48} & \best{77.82} & \best{77.14} & {\reqyes} & {\reqyes} & \textit{TFZS~w/o~Ext.} \\
\hline
\multicolumn{11}{c}{}\\[-6pt]
\hline
\multirow{9}{*}{\rotatebox[origin=c]{90}{Real3D-AD~\citep{ref22}}} & BTF~\citep{ref13}$^{\dagger}$ & 51.90 & 2.20 & 18.60 & 51.00 & 61.00 & 70.00 & {\reqyes} & {\reqno} & \multirow{5}{*}{\textit{Unsupervised}} \\
 & M3DM (MAE)~\citep{ref32}$^{\dagger}$ & 59.80 & 2.30 & 27.60 & 54.60 & 54.70 & 68.40 & {\reqyes} & {\reqno} &  \\
 & PC (MAE)~\citep{ref30}$^{\dagger}$ & 55.10 & 5.70 & 29.70 & 50.50 & 60.20 & 68.70 & {\reqyes} & {\reqno} &  \\
 & CPMF~\citep{ref05}$^{\dagger}$ & 75.90 & -- & -- & 58.60 & -- & -- & {\reqyes} & {\reqno} &  \\
 & Reg3DAD~\citep{ref22}$^{\dagger}$ & 69.50 & 11.50 & 30.60 & \second{69.00} & \second{70.90} & \second{73.10} & {\reqyes} & {\reqno} &  \\
\cline{2-11}
 & BTP~\citep{ref16}$^{\ddagger}$ & \second{80.61} & 12.96 & \second{50.39} & 65.32 & 67.95 & 71.76 & {\reqno} & {\reqno} & \textit{ZS} \\
\cline{2-11}
 & ZUMA~\citep{ref24}$^{\ddagger}$ & 62.33 & 2.14 & 20.23 & 53.45 & 55.33 & 68.21 & {\reqyes} & {\reqno} & \multirow{2}{*}{\textit{TFZS~w/~Ext.}} \\
 & MuSc-V2~\citep{ref19}$^{\ddagger}$ & 74.35 & \best{13.87} & 36.77 & 65.62 & 67.07 & 72.84 & {\reqyes} & {\reqno} &  \\
\cline{2-11}
 & \textbf{ASM}$^{\ddagger}$ & \best{84.15} & \second{13.06} & \best{56.08} & \best{71.28} & \best{73.03} & \best{75.01} & {\reqyes} & {\reqyes} & \textit{TFZS~w/o~Ext.} \\
\noalign{\hrule height 1.2pt}
\end{tabularx}
\par\vspace{3pt}
\begin{minipage}{\textwidth}
\footnotesize
$\dagger$: published results~\citep{ref16,ref17}. $\ddagger$: our evaluation (2,048 points). --: not reported. Red/blue: best/second-best shown. Local O-F1: maximum over thresholds. {\reqyes/\reqno: criterion met/not met. Training-Free: no task-specific parameter optimization. No Ext. Refs.: no reference banks, other instances, or text. Definitions: Sec.~4.2 and Supplementary Sec.~I.}
\end{minipage}
\ASMFloatBodyGap{8.34pt}
\end{table*}

 
\subsection{Computational Complexity}
Computing the full feature-distance matrix costs $O(M^2d)$. Gram construction and eigendecomposition cost $O(MK^2d+MK^3)$. Excluding the encoder and neighbor search, local modeling costs $O(K^2d+K^3)$ per query. With $K=16\ll d=384$, models use small neighborhood matrices, allowing queries to be batched or streamed. The supplement gives the derivation and component timings.
\section{Experiments}
\label{sec:experiments}
\subsection{Datasets and Evaluation}
We evaluate the complete test sets of Real3D-AD~\citep{ref22}, Anomaly-ShapeNet~\citep{ref17}, and IEC3D-AD~\citep{ref12}: 12/40/15 categories and 1,206/1,312/1,800 objects, totaling 67 categories and 4,318 objects. For locally evaluated methods, metrics are computed per category and then macro-averaged. Published baseline results and their aggregate scores follow the original reports. {Object-level detection uses AUROC, average precision (AP), and F1. Local results use maximum F1, while published F1 values follow the cited reports.} {Point-level localization uses AUROC, AP, and PRO.} For IEC3D-AD, we also report point-level AUPR and the ASM value is computed by trapezoidal integration of the precision--recall curve. Patch scores are averaged over overlapping patches for point evaluation. The default method and matched baselines use the same object-level anomaly score aggregation. Table~\ref{tab:benchmark-combined} jointly reports point-level localization and object-level detection on Anomaly-ShapeNet and Real3D-AD. Table~\ref{tab:benchmark-iec3d} reports point-level results on IEC3D-AD. Supplementary Tables~S1/S2 provide the category results for our local evaluations and available published breakdowns. Tables~S3/S4 provide IEC3D-AD results. Compact tables abbreviate the first two datasets as R3D and ASN, and AUROC as AUC. O and P denote object- and point-level results. {PC denotes \mbox{PatchCore in the comparison tables.}}

\Needspace*{6\baselineskip}
\subsection{Implementation and Comparison Protocols}
We use a frozen ULIP-2-pretrained Point-BERT encoder with 2,048 input points and 512 overlapping patches of 32 points.
Final-layer tokens undergo LayerNorm and FP32 $\ell_2$ normalization.
We fix $K=16$, $r=8$, $\epsilon=10^{-8}$, and $\tau=0.10$ across datasets.
Each object is processed independently, and point scores are \mbox{obtained by overlap averaging.}

{Tables~\ref{tab:benchmark-combined} and \ref{tab:benchmark-iec3d} mark training-free operation (TF) and no external references (No Ext. Refs.). \reqyes\ indicates satisfaction.}
{Training-Free excludes task-specific optimization, not generic pretraining or memory storage. External references include feature banks, other objects, and text.} A training-free method may still rely on external information at inference, so we report the two properties independently.
We report our own evaluations of BTP, ZUMA, MuSc-V2, and ASM, while published results retain their original protocols.
BTP uses a three-source one-vs-rest evaluation, with aggregation details in Supplementary Sec.~C.1.
Baseline sources {and detailed property definitions} are provided \mbox{in Supplementary Sec.~I.}

\subsection{Benchmark Results}
\paragraph{Real3D-AD and Anomaly-ShapeNet.}
{Table~\ref{tab:benchmark-combined} compares localization, detection, and the two property indicators on both benchmarks.} \method achieves the highest mean O- and P-AUROC among the compared TFZS methods on both benchmarks. Relative to the strongest TFZS~w/~Ext. comparator for each metric, its O-AUROC/P-AUROC gains are 5.66/9.80 points on Real3D-AD and 13.94/25.47 points on Anomaly-ShapeNet. Supplementary Tables~S1 and S2 provide available category breakdowns from the same evaluation records. {Unlike the normal-memory baselines, ASM builds support within the current test object instead of retaining target-category reference features. It also avoids the text cues used by ZUMA and the cross-object scoring used by MuSc-V2. {These results demonstrate ASM's effectiveness in point-level localization and object-level detection without task-specific training or external references.}}

{ASM reaches 31.65\% P-AP on Anomaly-ShapeNet, compared with 17.46\% for BTP. On Real3D-AD, its P-AP of 13.06\% is below MuSc-V2 (13.87\%) but above BTP (12.96\%). Localization is strong on diamond, gemstone, shell, and toffees, while fish and seahorse remain challenging. Strong results across bottles, bowls, caps, cups, and vases further highlight ASM's localization capability under a fixed configuration across all categories.}

\begin{table}[!t]
\TableRankingColorsOnly
\centering
\caption{Point-level anomaly localization on IEC3D-AD (\%).}
\label{tab:benchmark-iec3d}
\small
\ASMTableSpacing
\renewcommand{\arraystretch}{1.21}
\setlength{\arrayrulewidth}{0.4pt}
\begin{tabularx}{\columnwidth}{@{}l|>{\hsize=1.08\hsize\linewidth=\hsize\centering\arraybackslash}X>{\hsize=.92\hsize\linewidth=\hsize\centering\arraybackslash}X|cc@{}}
\noalign{\hrule height 1.2pt}
\multirow{2}{*}{Method} & \multicolumn{2}{c|}{Point-Level} & \multirow{2}{*}{{\makecell{Training-\\Free}}} & \multirow{2}{*}{{\makecell{No Ext.\\Refs.}}} \\
\cline{2-3}
& P-AUROC & P-AUPR & & \\
\hline
\multicolumn{5}{@{}c@{}}{\textit{Unsupervised}} \\
\hline
BTF (Raw)~\citep{ref13}$^{\dagger}$ & 70.17 & 2.82 & {\reqyes} & {\reqno} \\
BTF (FPFH)~\citep{ref13}$^{\dagger}$ & 53.40 & 0.68 & {\reqyes} & {\reqno} \\
M3DM (MAE)~\citep{ref32}$^{\dagger}$ & 55.65 & 1.79 & {\reqyes} & {\reqno} \\
M3DM (BERT)~\citep{ref32}$^{\dagger}$ & 54.17 & 1.66 & {\reqyes} & {\reqno} \\
PC (MAE)~\citep{ref30}$^{\dagger}$ & 66.97 & 4.99 & {\reqyes} & {\reqno} \\
PC (FPFH)~\citep{ref30}$^{\dagger}$ & 64.26 & 4.50 & {\reqyes} & {\reqno} \\
PC (FPFH+Raw)~\citep{ref30}$^{\dagger}$ & 66.86 & 5.95 & {\reqyes} & {\reqno} \\
Reg3DAD~\citep{ref22}$^{\dagger}$ & 69.89 & 7.78 & {\reqyes} & {\reqno} \\
IMRNet~\citep{ref17}$^{\dagger}$ & 71.64 & 8.45 & {\reqno} & {\reqno} \\
R3D-AD~\citep{ref43}$^{\dagger}$ & 69.57 & 8.12 & {\reqno} & {\reqyes} \\
GMANet~\citep{ref12}$^{\dagger}$ & \second{72.52} & \second{10.89} & {\reqno} & {\reqyes} \\
\hline
\multicolumn{5}{@{}c@{}}{\textit{TFZS~w/o~Ext.}} \\
\hline
\textbf{ASM}$^{\ddagger}$ & \best{92.37} & \best{15.41} & {\reqyes} & {\reqyes} \\
\noalign{\hrule height 1.2pt}
\end{tabularx}
\par\vspace{2pt}
\begin{minipage}{\columnwidth}
\footnotesize
$\dagger$: published five-run means~\citep[Table~V]{ref12}. $\ddagger$: local evaluation (2,048 points). Red/blue: best/second-best values shown. {Flags follow Table~\ref{tab:benchmark-combined}.}
\end{minipage}
\ASMFloatBodyGap{9.94pt}
\end{table}

\paragraph{IEC3D-AD.}
{Table~\ref{tab:benchmark-iec3d} reports mean P-AUROC and P-AUPR alongside the same indicators of training-free operation and reference-free inference.} ASM achieves 92.37\% and 15.41\%, respectively, compared with 72.52\% and 10.89\% for GMANet. Supplementary Table~S3 gives the category-level P-AUROC results, where ASM ranks first in 14 of 15 categories. Its category scores range from 87.60 to 99.60, with only butterfly bolt ranking second to IMRNet. The improvement therefore covers fasteners with different shapes and defect patterns, including rings, nuts, bolts, studs, and nails. Supplementary Table~S4 provides the corresponding object-level results. These results demonstrate effective defect localization \mbox{across diverse fastener geometries.}

\begin{figure*}[!t]
\centering
\begin{lrbox}{\ASMsupportbox}
\begin{minipage}[t]{\columnwidth}
\vspace{0pt}
\ASMTableCaptionSetup
\TableRankingColorsOnly
\centering
\caption{Intra-object support ablation (P-AUROC / {O-AUROC}, \%).}
\label{tab:reference-ablation}
\small
\ASMTableSpacing
\begin{tabular*}{\linewidth}{@{\extracolsep{\fill}}lcc@{}}
\toprule
Support & R3D & ASN \\
\midrule
Random & {71.53$\pm$0.48/62.75$\pm$0.70} & {77.66$\pm$0.30/56.43$\pm$1.01} \\
Global affine & 75.52 / {65.41} & 83.25 / {61.83} \\
Spatial KNN & 82.84 / {70.92} & 89.50 / {69.62} \\
Feature KNN & \textbf{84.15} / {\textbf{71.28}} & \textbf{91.06} / {\textbf{71.48}} \\
\bottomrule
\end{tabular*}
\par\vspace{2pt}
{\scriptsize\raggedright{Random: mean $\pm$ sample standard deviation across three seeds.}\par}

\end{minipage}
\end{lrbox}
\begin{lrbox}{\ASMbackbonebox}
\begin{minipage}[t]{\columnwidth}
\vspace{0pt}
\ASMTableCaptionSetup
\TableRankingColorsOnly
\centering
\caption{Frozen-encoder ablation (P-AUROC / {O-AUROC}, \%).}
\label{tab:backbone}
\small
\ASMTableSpacing
\begin{adjustbox}{max width=\columnwidth}
\begin{tabular*}{\linewidth}{@{\extracolsep{\fill}}llcc@{}}
\toprule
Backbone & {Scorer} & R3D & ASN \\
\midrule
\multirow{2}{*}{ULIP2-PB} & {LM (sq.)} & {80.80 / 70.53} & {88.01 / 67.40} \\
& {ASM} & \textbf{84.15} / {\textbf{71.28}} & \textbf{91.06} / {\textbf{71.48}} \\
\midrule
\multirow{2}{*}{Point-BERT} & {LM (sq.)} & {65.60 / 56.33} & {69.68 / 57.81} \\
& {ASM} & \textbf{70.82} / {\textbf{59.24}} & \textbf{76.88} / {\textbf{58.97}} \\
\midrule
\multirow{2}{*}{Point-MAE} & {LM (sq.)} & {67.16 / 63.11} & {73.92 / 68.02} \\
& {ASM} & \textbf{78.56} / {\textbf{65.91}} & \textbf{81.82} / {\textbf{71.04}} \\
\bottomrule
\end{tabular*}
\end{adjustbox}

\end{minipage}
\end{lrbox}
\begin{lrbox}{\ASMscoringbox}
\begin{minipage}[t]{\columnwidth}
\vspace{0pt}
\ASMTableCaptionSetup
\TableRankingColorsOnly
\centering
\caption{Scoring-rule ablation (P-AUROC / O-AUROC, \%).}
\label{tab:score-audit}
\small
\ASMTableSpacing
\begin{adjustbox}{max width=\columnwidth}
\begin{tabular*}{\linewidth}{@{\extracolsep{\fill}}lcc@{}}
\toprule
Scorer & R3D & ASN \\
\midrule
1-NN (sq.) & 79.80 / 68.14 & 88.28 / 67.60 \\
KNN mean (sq.) & 77.32 / 68.53 & 84.83 / 63.61 \\
Local mean (sq.) & 80.80 / 70.53 & 88.01 / 67.40 \\
Global affine (sq.) & 75.86 / 65.44 & 83.85 / 61.84 \\
ASM (sq.) & 84.26 / 71.14 & 91.23 / 71.66 \\
ASM & 84.15 / 71.28 & 91.06 / 71.48 \\
\bottomrule
\end{tabular*}
\end{adjustbox}
\end{minipage}
\end{lrbox}
\begin{lrbox}{\ASMlatencybox}
\begin{minipage}[t]{\columnwidth}
\vspace{0pt}
\ASMTableCaptionSetup
\TableRankingColorsOnly
\centering
\caption{Inference latency and peak GPU memory across methods.}
\label{tab:deployment-speed}
\small
\ASMTableSpacing
\begin{adjustbox}{max width=\columnwidth}
\begin{tabular*}{\linewidth}{@{\extracolsep{\fill}}lccc@{}}
\toprule
Method & R3D & ASN & Memory \\
\midrule
BTP & 22.74 / 43.97 & 19.84 / 50.40 & 10073.60 \\
ZUMA & 1799.12 / 0.56 & 1698.98 / 0.59 & 2435.70 \\
MuSc-V2 & 251.50 / 3.98 & 383.98 / 2.60 & 820.40 \\
\textbf{ASM} & \textbf{9.91 / 100.93} & \textbf{9.73 / 102.74} & \textbf{606.40} \\
\bottomrule
\end{tabular*}
\end{adjustbox}
\end{minipage}
\end{lrbox}
\noindent\begin{minipage}[t]{\columnwidth}
\vspace{0pt}
\usebox{\ASMsupportbox}
\par\vspace{7pt}
\usebox{\ASMscoringbox}
\end{minipage}\hfill
\begin{minipage}[t]{\columnwidth}
\vspace{0pt}
\usebox{\ASMbackbonebox}
\par\vspace{10.45pt}
\usebox{\ASMlatencybox}
\end{minipage}

\par\vspace{7pt}
\captionsetup{skip=3pt}
\includegraphics[width=\textwidth,trim=0 22bp 0 30bp,clip]{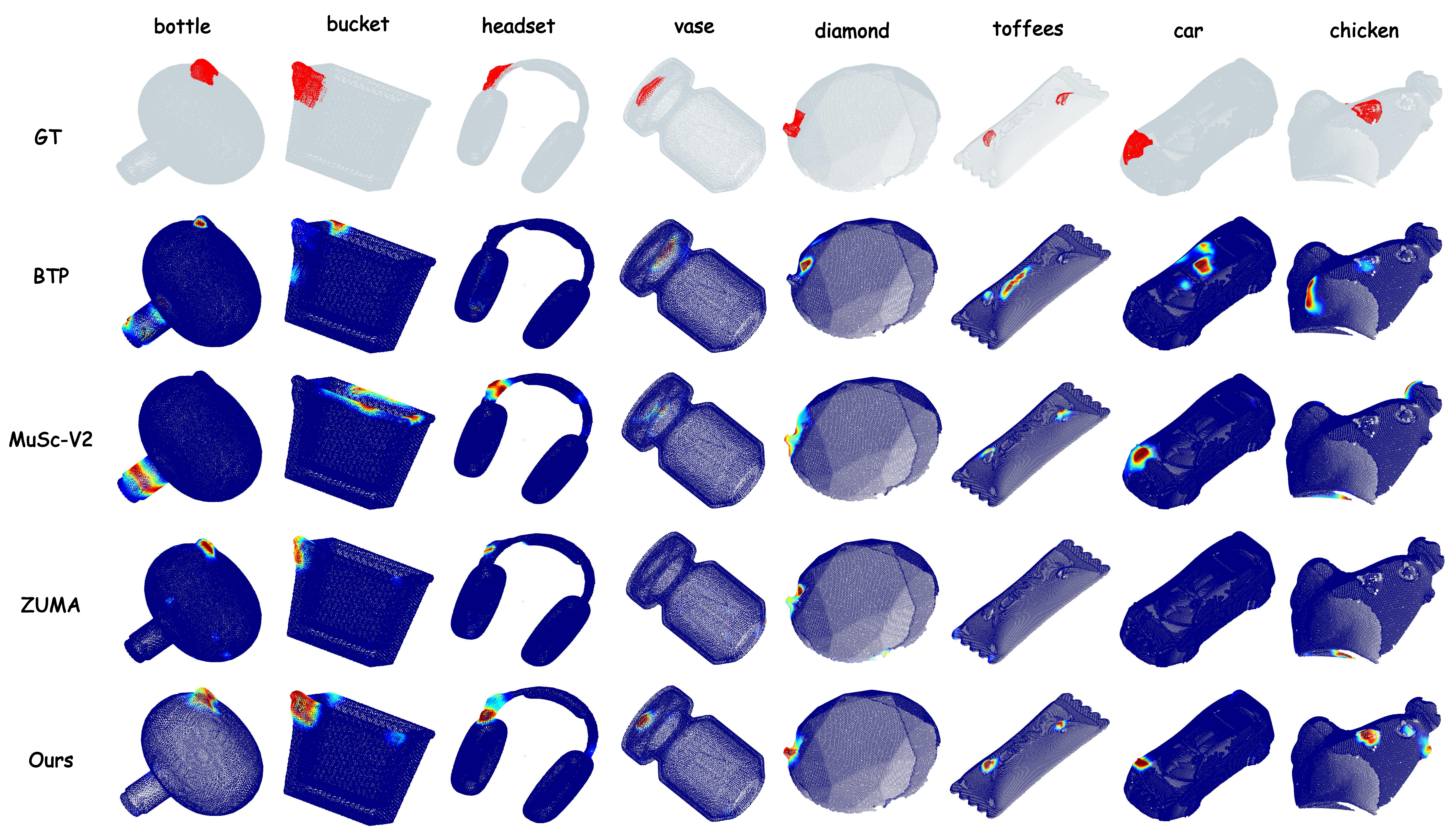}
\caption{Qualitative comparison on Anomaly-ShapeNet
(left four columns) and Real3D-AD (right four columns).
Red GT points indicate defects, and warmer prediction
colors indicate higher anomaly scores.}
\label{fig:qualitative}

\par\ASMFloatBodyGap{9.34pt}
\end{figure*}

\subsection{Ablation Studies}
\label{sec:ablation}
\paragraph{Support Construction.}
Table~\ref{tab:reference-ablation} fixes the encoder, residual definition, and score aggregation while varying intra-object support. {Feature KNN improves P-AUROC over spatial KNN by 1.31/1.55 points and O-AUROC from 70.92/69.62 to 71.28/71.48 on Real3D-AD/Anomaly-ShapeNet. It also outperforms global support, despite the latter using all patches. Random support reports the mean and sample standard deviation over three seeds. Supplementary Table~S8 provides P-AP, PRO, and per-seed results. These results demonstrate the effectiveness of ASM's query-specific feature support for \mbox{both localization and detection.}}

\paragraph{Different Pretrained Encoders.}
Table~\ref{tab:backbone} compares ASM with squared {local-mean (LM)} scoring on identical features from frozen ULIP2-PB, Point-BERT, and Point-MAE~\citep{ref26}. {Both scoring rules are listed for each encoder. Across all six dataset--encoder combinations, ASM achieves higher P-AUROC and O-AUROC than the local-mean baseline. Supplementary Table~S9 provides the complete results. These results demonstrate the effectiveness of ASM for both localization and detection across \mbox{the three frozen encoders.}}

\paragraph{Effect of Affine Modeling.}
Table~\ref{tab:score-audit} compares scoring rules with frozen features, neighborhoods, and aggregation fixed. Sq. denotes squared distances or residual norms before aggregation. Local mean (sq.) uses the support center, whereas ASM (sq.) also models variation directions. Under this matched convention, P-AUROC increases from 80.80/88.01 to 84.26/91.23 on Real3D-AD/Anomaly-ShapeNet, gains of 3.45/3.22 points. O-AUROC increases from 70.53/67.40 to 71.14/71.66. Default ASM uses the residual norm and obtains 84.15/91.06 P-AUROC. Supplementary Sec.~D gives the P-AP comparisons and category-paired bootstrap intervals. These gains support local affine modeling as an effective approach to anomaly localization.

\subsection{Sensitivity and Qualitative Comparison}
\paragraph{Parameter Sensitivity.}
Localization varies little across the tested settings with support size $K\in[12,24]$ and maximum rank $r\in[8,14]$. We use $K=16$ and $r=8$ for all categories and datasets, balancing local fitting and computational cost. Supplementary Fig.~S1 and Tables~S11/S12 provide the complete sweeps. Table~S13 \mbox{compares object-level aggregation rules.}

\paragraph{Qualitative Comparison.}
Figure~\ref{fig:qualitative} compares ASM, BTP, MuSc-V2, and ZUMA on eight examples. ASM localizes annotated defects in bottle, bucket, headset, vase, diamond, car, and chicken, including both separated defects on toffees. BTP responds weakly to the headset defect, while ZUMA responds weakly on toffees and car. MuSc-V2 gives strong responses in some unannotated regions of bottle and chicken. These examples highlight ASM's ability to localize defects across different object geometries.

\subsection{{Affine Residual Analysis}}
\label{sec:mechanism-analysis}
\begin{figure}[!t]
\centering
\includegraphics[width=0.962\linewidth]{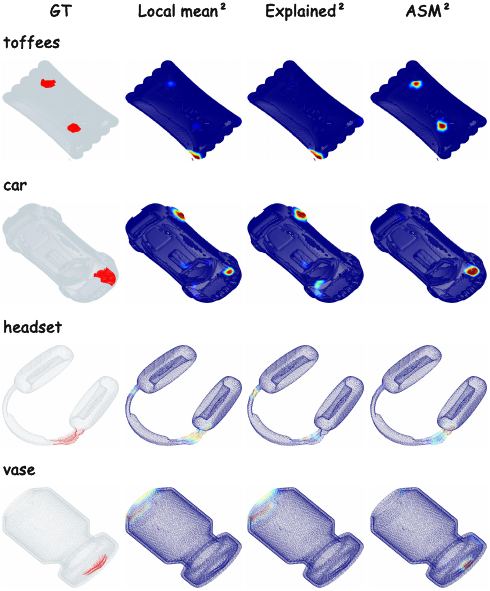}
\caption{{\textbf{Spatial responses under affine self-modeling.} Four examples from Real3D-AD (top) and Anomaly-ShapeNet (bottom). Columns show GT, total deviation energy (Local mean$^2$), model-explained energy (Explained$^2$), and residual energy (ASM$^2$). {Energy denotes the squared feature-space quantities in the affine decomposition, computed at the patch level before overlap averaging.} Independent color scales show relative response locations, \mbox{not comparable absolute magnitudes.}}}
\label{fig:asm-mechanism-examples}
\ASMFloatBodyGap{4.86pt}
\end{figure}

\paragraph{{Spatial Response Patterns.}}
{Figure~\ref{fig:asm-mechanism-examples} illustrates how local affine modeling helps distinguish normal feature variation from anomalous deviations. With frozen features and support sets fixed, off-defect responses in the local-mean maps also appear in the model-explained component, whereas the affine residual responses are more concentrated around annotated defects. This pattern is particularly clear in toffees, car, and vase. These observations support the ability of the local affine subspaces to account for normal structural variation while retaining anomalous deviations in the residuals.}

\begin{figure}[!t]
\centering
\includegraphics[width=0.968\linewidth,trim=3.1497bp 0 3.1497bp 0,clip]{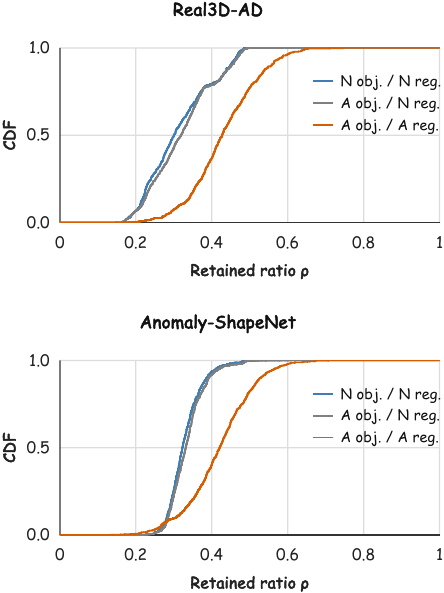}
\caption{{\textbf{Retained-energy distributions over the complete test sets.} Observations are within-object regional medians of pointwise residual-to-total energy ratios. N/A denote normal/anomalous, and obj./reg. denote object/region. All 1,206 Real3D-AD and 1,312 Anomaly-ShapeNet test objects contribute equally within each group. Statistics precede display normalization.}}
\label{fig:asm-retention-full-test}
\ASMFloatBodyGap{6.41pt}
\end{figure}

\paragraph{{Residual Energy Retention.}}
{Figure~\ref{fig:asm-retention-full-test} extends the analysis to all 1,206 Real3D-AD and 1,312 Anomaly-ShapeNet test objects. From raw point-level energies, we compute $\rho_p=E_{\mathrm{res},p}/(E_{\mathrm{total},p}+10^{-12})$ and take its median within each normal or anomalous region, with one observation per object and region type. Normal regions are grouped separately for normal and anomalous objects rather than pooled into a single distribution. Equal weighting within each group also prevents objects with larger regions from dominating these statistics. For anomalous objects, both regional summaries come from the same input cloud. Anomalous regions generally retain a higher proportion, whereas normal-region distributions are close across object groups. This supports preferential retention of anomalous deviations, with overlap between the distributions. Supplementary Sec.~J \mbox{gives the aggregation details.}}


\subsection{Efficiency}
\label{sec:efficiency}
\paragraph{Deployment Cost.}
Table~\ref{tab:deployment-speed} reports latency (ms/object / FPS) and peak GPU memory (MiB) for 2,048-point inputs on an NVIDIA GeForce RTX 5090 D. ASM takes 9.91/9.73 ms on Real3D-AD/Anomaly-ShapeNet and uses 606.40 MiB. Timing excludes model loading, training, and metric computation. ASM processes each cloud independently, so feature extraction and affine fitting are not amortized over additional test objects. Method-specific scopes and evaluation cohorts are detailed in Supplementary Sec.~H. These measurements demonstrate the low inference cost of \mbox{ASM for single-object inspection} without caching features across test objects.

\paragraph{Cost of Local Modeling.}
Supplementary Sec.~H profiles 104 objects over three CUDA-synchronized repetitions. Table~S18 reports median costs of 0.537\,ms for local modeling, 4.958\,ms for feature extraction, and 7.72\,ms for the prepared-input pipeline. Table~S17 separately times scorers on cached features. {Local modeling remains sub-millisecond, with the frozen backbone accounting for most computation.}

\par\begingroup\setlength{\parskip}{0pt}\vspace{6pt}\indent
Tables~\ref{tab:benchmark-combined} and~\ref{tab:deployment-speed} show ASM's accuracy and efficiency.\par\endgroup

\section{Limitations}
ASM depends on the quality of its frozen features and intra-object support. Unique normal structures may have few suitable neighbors, while repeated defects may support one another. Defects aligned with retained variation directions can receive low scores. Point and patch resolution limit the detection of small defects. The current formulation focuses on local feature variation, leaving category-level logical constraints and cross-object score calibration outside its scope.
\section{Conclusion}
We presented ASM for training-free zero-shot point cloud anomaly detection using query-specific affine models within each object. By modeling the feature variation of similar intra-object regions, ASM uses affine residuals to assess local deviations without normal reference objects, other test instances, or text prompts. ASM achieves 84.15\%, 91.06\%, and 92.37\% point-level AUROC on Real3D-AD, Anomaly-ShapeNet, and IEC3D-AD. Matched scoring comparisons and cross-encoder experiments support local variation modeling. Small Gram matrices support efficient single-object inference \mbox{without additional trainable parameters.} Together, these findings support the view that a test cloud can provide not only comparison features, but also a local model of how related regions vary.

Future work will investigate more reliable support selection, adaptive local-rank estimation, and multi-scale feature modeling to better handle sparse or defect-contaminated supports and improve sensitivity to small defects. We will also study object-level score aggregation and within-object calibration to strengthen detection across diverse geometries while preserving training-free, single-object inference \mbox{without external reference data.}
 \clearpage
\section*{LLM Usage Statement}
Chat GPT was used to assist with manuscript
writing and language editing. Specifically, it helped refine
the wording across sections, improve readability, and ensure
clarity. These tasks included sentence rephrasing, grammatical
proofreading, and improving the logical flow of the manuscript.
\nocite{gsclip2026,ref41,ref45}
 {\small\bibliographystyle{ieeenat_fullname}\bibliography{main}}
\end{document}

%% file: preamble.tex
\usepackage[T1]{fontenc}
\usepackage{mathtools,bm,multirow,makecell,tabularx,adjustbox}
\usepackage{microtype,currfile}
\usepackage{enumitem}
\usepackage{needspace}
\definecolor{bestred}{RGB}{190,35,35}
\definecolor{secondblue}{RGB}{30,85,175}
\newcommand{\best}[1]{\textcolor{bestred}{\textbf{#1}}}
\newcommand{\second}[1]{\textcolor{secondblue}{#1}}
\newcommand{\method}{ASM\xspace}

\DeclareCaptionLabelFormat{parta}{#1~#2(a)}
\DeclareCaptionLabelFormat{partb}{#1~#2(b)}
\DeclareCaptionLabelFormat{partc}{#1~#2(c)}
\setlist[itemize]{leftmargin=*,itemsep=2pt,topsep=3pt}


\newcommand{\TableRankingColorsOnly}{%
  \color{black}%
  \hypersetup{allcolors=black}%
}

\newsavebox{\ASMsupportbox}
\newsavebox{\ASMbackbonebox}
\newsavebox{\ASMscoringbox}
\newsavebox{\ASMlatencybox}

\makeatletter
\newcommand{\ASMTableCaptionSetup}{%
  \def\@captype{table}%
  \captionsetup{type=table,skip=4pt}%
}
\makeatother

\newcommand{\ASMTableSpacing}{%
  \renewcommand{\arraystretch}{1.10}%
  \setlength{\tabcolsep}{2pt}%
}